\documentclass[10pt,twocolumn,letterpaper]{article}

\usepackage{cvpr}              

\usepackage[dvipsnames]{xcolor}

\definecolor{cvprblue}{rgb}{0.21,0.49,0.74}
\usepackage[pagebackref,breaklinks,colorlinks,citecolor=cvprblue]{hyperref}

\usepackage{graphicx}
\usepackage{amsmath}
\usepackage{amssymb}
\usepackage{booktabs}
\usepackage{tabularx,booktabs}
\usepackage{enumitem}
\usepackage[font=small,skip=1pt]{caption}
\usepackage{multirow}
\usepackage{nicematrix}

\def\paperID{8979} 
\def\confName{CVPR}
\def\confYear{2024}

\title{Mixed-Query Transformer: A Unified Image Segmentation Architecture}

\author{Pei Wang$^{1}$, Zhaowei Cai$^1$, Hao Yang$^1$, Ashwin Swaminathan$^1$, R. Manmatha$^2$, Stefano Soatto$^2$ \\ [.5ex]
$^1$AGI \qquad $^2$AWS AI Labs \\
\tt\small \{pwwng,zhaoweic,haoyng,swashwin,manmatha,soattos\}@amazon.com
}

\begin{document}
\maketitle

\begin{abstract}
Existing unified image segmentation models either employ a unified architecture across multiple tasks but use separate weights tailored to each dataset, or apply a single set of weights to multiple datasets but are limited to a single task. In this paper, we introduce the Mixed-Query Transformer (MQ-Former), a unified architecture for multi-task and multi-dataset image segmentation using a single set of weights. To enable this, we propose a mixed query strategy, which can effectively and dynamically accommodate different types of objects without heuristic designs. In addition, the unified architecture allows us to use data augmentation with synthetic masks and captions to further improve model generalization. Experiments demonstrate that MQ-Former can not only effectively handle multiple segmentation datasets and tasks compared to specialized state-of-the-art models with competitive performance, but also generalize better to open-set segmentation tasks, evidenced by over 7 points higher performance than the prior art on the open-vocabulary SeginW benchmark.

\end{abstract}

\section{Introduction}
\label{sec:intro}

\begin{figure*}[t]
\setlength{\abovecaptionskip}{-2.0pt}
\setlength{\tabcolsep}{2pt}
\begin{center}
  \includegraphics[width=1.0\linewidth]{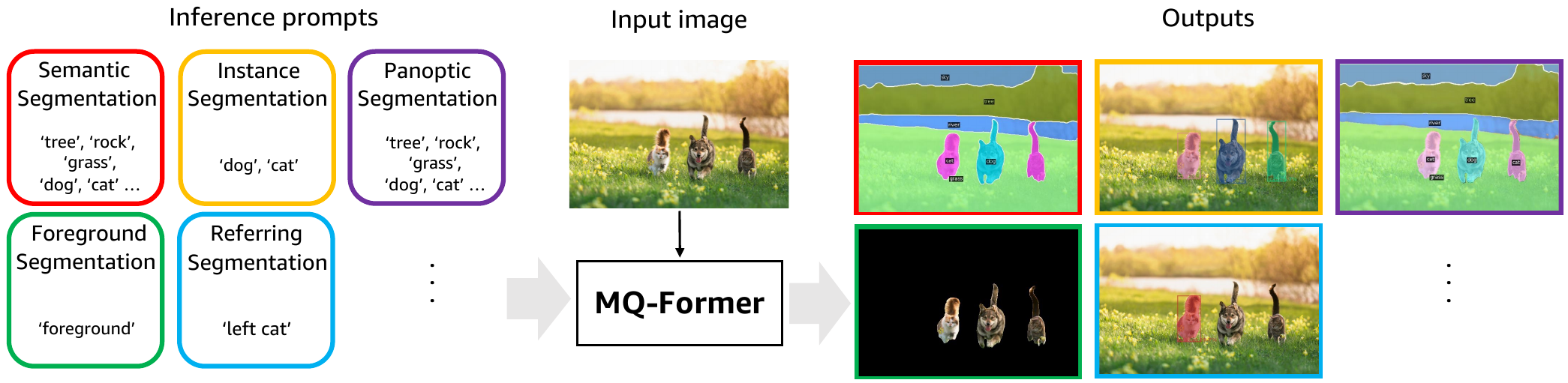}
\end{center}
  \caption{The supported tasks of MQ-Former. Within one {\bf single} training configuration, MQ-Former supports jointly training on multiple segmentation tasks and datasets, and inference with open-vocabulary setting for multiple tasks.
  }
\label{fig:teaser}
\end{figure*}

Image segmentation is an important computer vision research direction with the goal of partitioning an image into discrete groups of pixels. It has various branches, including semantic segmentation, instance segmentation, panoptic segmentation, foreground/background or saliency segmentation, referring segmentation, etc. The objective of a universal image segmentation model is to exhibit robust generalization capabilities, performing effectively in real-world open-world applications. To achieve that, a straightforward solution is to train models on a diverse range of data and tasks within a unified architecture, such that the knowledge across different tasks and datasets can be integrated. However, numerous prior works were independently explored on each task or dataset~\cite{he2017mask,chen2017deeplab,chen2019hybrid,ronneberger2015u,long2015fully,xiong2019upsnet,cheng2020panoptic}. These specialized models, while contributing significantly to their respective tasks, usually fail to address real-world scenarios, where versatility and adaptability are critical.


Several recent transformer-based approaches have explored unified segmentation frameworks, seeking to address the joint training of multiple tasks and datasets~\cite{cheng2022masked,li2023mask,jain2023oneformer,kim2022learning,cheng2021per, zhou2023lmseg,zou2023generalized,zhang2023simple}. However, these existing works possess certain inherent limitations, far from the unification of \emph{more} segmentation tasks and datasets. \cite{cheng2022masked,li2023mask,cheng2021per} require architectural modifications to address different tasks. \cite{jain2023oneformer} can handle multiple tasks but relies on separate weights for each dataset, such as COCO \cite{lin2014microsoft} or ADE \cite{zhou2017scene}. \cite{zou2023generalized,zhang2023simple} are trainable only on the panoptic segmentation task, which requires more expensive annotation compared to semantic or instance segmentation.
None of these prior frameworks has fully leveraged the diverse information across segmentation tasks and datasets. A key factor preventing the unification relates to object query design, which is a core architectural component in transformer-based segmentation models. In the literature, learnable queries~\cite{carion2020end} have proven effective for semantic (stuff) segmentation~\cite{zou2023generalized,li2023mask}, but conditional queries~\cite{zhu2020deformable} (similar to proposal detection~\cite{ren2015faster}) excel at instance (thing) segmentation~\cite{li2023mask,zhang2023simple}\footnote{The term ``thing'' (referring to countable objects, usually in the foreground) and ``stuff'' (referring non-object, uncountable elements, often in the background) are frequently employed to make a distinction between objects with clearly defined geometry and quantifiability, such as people, dogs, and surfaces or areas lacking a fixed geometry, primarily recognized by their texture or material, like sky, road~\cite{kirillov2019panoptic}.}. Although OpenSeeD~\cite{zhang2023simple} proposes a separate query mechanism in order to separate learnable (conditional) queries to match stuff (thing) objects, this design requires first classifying all targets into stuff/thing and then applying the corresponding learnable/conditional queries at both training and inference. Such an approach is heuristic and impractical, since there is no clear delineation between things and stuff in general, and this categorical information is unavailable in most of public datasets, \eg, Visual Genome~\cite{krishna2017visual}, especially in the open-vocabulary segmentation setting. These challenges impede existing segmentation architectures from effectively handling diverse datasets tailored for specific tasks, resulting in suboptimal performance in open-vocabulary segmentation scenarios.

In this work, we introduce a new unified segmentation architecture capable of effectively handling a diverse array of tasks and datasets, demonstrating robust performance in real-world open-set scenarios.
We first propose a novel object query approach termed {\it mixed query}. It combines learnable and conditional queries, with the mixed queries being matched to thing and stuff objects automatically via Hungarian matching. In contrast to prior query designs, 1) mixed query can accommodate both thing and stuff objects effectively; and 2) there is no heuristic discrimination between thing and stuff objects, since the Hungarian matching process will determine the optimal matching between objects and queries \emph{automatically}. This design facilitates universal queries for handling diverse object types. With the mixed query mechanism, we present a unified segmentation architecture, mixed query transformer (MQ-Former), which can be trained and evaluated on any segmentation task and dataset, as shown in Figure \ref{fig:teaser}, without the constraint of using only panoptic segmentation annotations as in \cite{zhang2023simple,zou2023generalized} and extra thing/stuff annotations as in \cite{zhang2023simple}.
This property additionally enables MQ-Former to leverage more existing segmentation datasets for performance improvement, \eg the data of referring \cite{kazemzadeh2014referitgame,yu2016modeling} and foreground/background \cite{ChengPAMI,mansilla2016oriented,liew2021deep,xie2022pyramid} segmentation, etc. 

As the core benefit of the unified design of MQ-Former, we are enticed to use more diverse segmentation data to further enhance the model generalization. However, human annotation for segmentation is usually expensive, \eg, requiring a few minutes to annotate a single COCO image.  
To circumvent this data limitation, we propose to harness synthetic data, \ie, synthetic segmentation masks for pixel-level segmentation and synthetic segment captions for open-vocabulary semantic alignment. This is feasible as some recent models can already generate impressive synthetic segmentation masks~\cite{kirillov2023segment,ke2023segment} and object-level captions~\cite{wang2022ofa,li2023blip}, and the synthetic data has been proven helpful for model improvement~\cite{cho2023open,gao2022open}. Given the low cost of generating synthetic data, we can readily accumulate a large-scale segmentation training set to enhance our MQ-Former. The incorporation of synthetic data not only addresses the challenge of data scarcity, but also augments model robustness and semantic understanding capabilities. In our experiments, with the help of mixed query and synthetic data training, our MQ-Former surpasses the state-of-the-art on many open-set segmentation benchmarks. For instance, on the in-the-wild SeginW benchmark~\cite{zou2023generalized}, MQ-Former outperforms the state-of-the-art by over 7 points. These advancements constitute a substantial stride towards a unified and well-generalized image segmentation model.

Overall, this paper has three major contributions. First, we design a novel mixed query mechanism which can effectively accommodate various segmentation tasks, without heuristic discrimination of thing and stuff classes. Second, with mixed query, we propose a unified segmentation architecture, MQ-Former, and it is the first work to have multi-task and multi-dataset training, to the best of our knowledge, and has shown successes. Third, in this unified MQ-Former, we have demonstrated the effectiveness of using synthetic data for improving overall segmentation performance, showing the state-of-the-art results on multiple open-set segmentation benchmarks.

\section{Related Work}
\label{sec:rela}

\noindent \textbf{Generic segmentation} Given an input image, the goal of image segmentation is to output a group of masks with class predictions. According to the scope of class labels and masks, image segmentation can be divided into three major tasks, semantic, instance and panoptic segmentation~\cite{li2023transformer}. In the past, many task or dataset specialized models have been proposed, and they can be trained and do inference only on a single task and dataset, including Mask R-CNN~\cite{he2017mask}, Cascade Mask R-CNN \cite{cai2019cascade}, HTC~\cite{chen2019hybrid} on instance segmentation, FCN~\cite{long2015fully}, U-Net~\cite{ronneberger2015u}, DeepLab~\cite{chen2017deeplab} on semantic segmentation, UPSnet~\cite{xiong2019upsnet}, Panoptic-DeepLab~\cite{cheng2020panoptic} on panoptic segmentation. 

\noindent \textbf{Unified segmentation models} The unified segmentation model has been explored, and it is a challenging task. Early works on unified models either support training on multiple tasks but distinct weights for different datasets~\cite{jain2023oneformer,zhang2021k,qin2023freeseg} or training on multiple datasets but different weights for each task~\cite{kim2022learning,lambert2020mseg, zhou2023lmseg}. Recently, there are several attempts to train a single model across datasets and tasks but the unification is not fully achieved yet. In the series of works of MaskFormer \cite{cheng2021per,cheng2022masked,li2023mask}, although the core architecture is the same across semantic, instance and panoptic segmentation tasks, the architecture and configuration still require slight modifications for different tasks. For example, in Mask DINO~\cite{li2023mask}, 
one-stage encoder-decoder architecture performs well on semantic segmentation but poorly on instance and panoptic segmentation. The behavior of two-stage counterpart is reversed.
This paradox impedes the joint training across different tasks without performance degradation. X-Decoder and DaTaSeg try to address this problem with a sub-optimal solution, by using a decoder-only one-stage architecture, but its instance segmentation performance is decreased~\cite{zou2023generalized,gu2024dataseg}. Other works use different queries or decoders for specific entities or tasks~\cite{zhang2023simple,rana2023dynamite,athar2023tarvis,wang2024hierarchical}. OpenSeeD~\cite{zhang2023simple} uses two sets of queries for stuff and thing classes separately. However, this introduces extra cost and uncertainty for distinguishing stuff/thing during data annotation and inference. Most of existing datasets lack such annotation information~\cite{krishna2017visual,kazemzadeh2014referitgame,yu2016modeling} and there is no clear delineation between things and stuff, making it impractical to annotate especially for open-vocabulary datasets.
To the best of our knowledge, there is no unified model yet that supports training on multiple datasets and tasks simultaneously, and has shown successes. In this work, the proposed MQ-Former tries to address this challenge.

\noindent\textbf{Using synthetic data for stronger model} \cite{cho2023open} uses an 
image captioning model to generate pseudo captions on the cropped object regions for object detection, but it neglects the context information during the object caption generation. Pseudo bounding boxes are also leveraged to expand the training data size~\cite{gao2022open}. For image segmentation, PseudoSeg~\cite{zou2020pseudoseg} designs a one-stage framework to generate pseudo masks from unlabeled data or image-level labeled data for semantic segmentation. Another line producing and applying pseudo labels to improve the model is under the teacher-student semi-supervised learning framework~\cite{chen2021semi,wang2022semi,liu2022perturbed}. OpenSeeD~\cite{zhang2023simple} also uses a pseudo mask generator decoding from bounding boxes during training. However, we argue that all these on-the-fly pseudo data generation methods will increase the training cost. In our work, inspired by the recent segmentation models that can generate high-quality mask predictions~\cite{kirillov2023segment,ke2023segment} and have been shown to be a good pseudo label generator~\cite{jiang2023segment,chen2023segment}, we generate the synthetic data offline, which will be used during training with no difference from ground truth.

\section{Method}
\label{sec:method}

In this section, we first provide an overview of the MQ-Former architecture. We then introduce the novel mixed query mechanism as a key component of the architecture. Next, we discuss how the mechanism can handle multi-task and multi-dataset joint training. Finally, we describe enhancements to the MQ-Former using synthetic data to train more robust models.

\subsection{MQ-Former Architecture}

\begin{figure}[t]
\setlength{\abovecaptionskip}{-2.0pt}
\setlength{\tabcolsep}{2pt}
\begin{center}
  \includegraphics[width=1.0\linewidth]{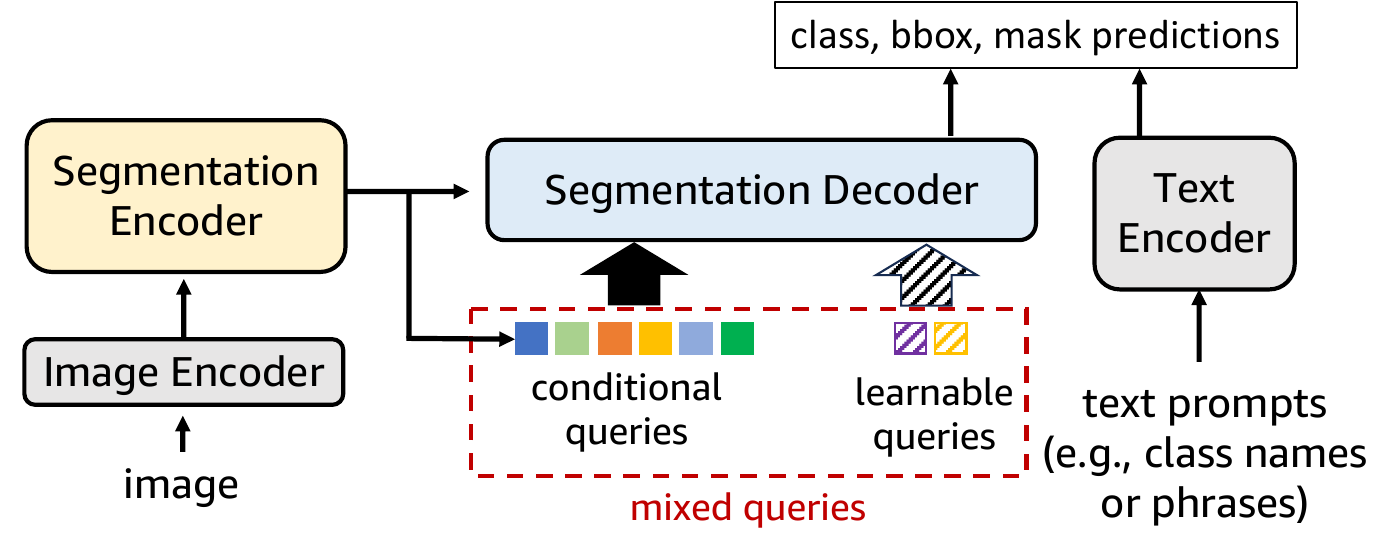}
\end{center}
  \caption{The overview of MQ-Former architecture. The model takes an image and a list of textual language prompts as input and outputs their corresponding localized segment masks.}
\label{fig:arch}
\end{figure}

Figure \ref{fig:arch} shows the architecture of the proposed MQ-Former. It has four major components, image and text encoder, and segmentation encoder and decoder. The image encoder encodes an input image to multi-scale image features, and the text encoder encodes the text query to obtain its semantic embedding. The multi-scale image features are forwarded to the segmentation encoder for further refinement. Next, the segmentation decoder takes numbers of object queries and cross-attends the refined image features to predict the final class, bounding box (bbox), and segment mask.

\subsection{Object Query Strategies}

\begin{figure*}[t]
\setlength{\abovecaptionskip}{-2.0pt}
\setlength{\tabcolsep}{2pt}
\begin{center}
  \includegraphics[width=1.0\linewidth]{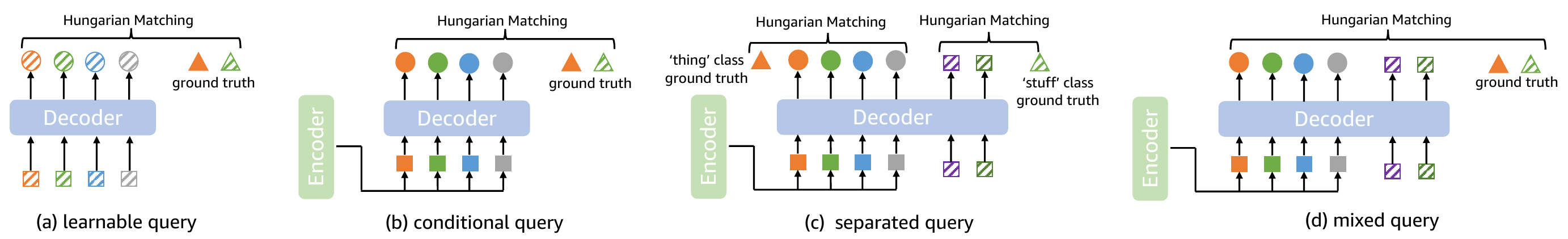}
\end{center}
  \caption{The comparison of different query strategies. Square with diagonal slashes: learnable query; solid square: conditional query; circle with slashes: query embedding of learnable queries; solid circle: query embedding of conditional queries; triangle with slashes: ground truth of stuff class; solid triangle: ground truth of thing classes. (a) learnable query is learned from scratch. (b) conditional query is derived and selected from encoder. (c) separated query consists of both learnable and conditional queries which are associated with stuff and thing classes respectively. (d) mixed query also consists of both types of queries but does not impose thing/stuff distinction.}
\label{fig:comp_query_stra}
\end{figure*}

Object query is a key component in transformer-based object detection and segmentation models,
and has attracted much attention from the community~\cite{liu2022dab,li2022dn,zhu2020deformable,zhang2022dino,wang2022anchor,meng2021conditional}. In this section, we first review three common object query strategies in segmentation tasks and then introduce our new mixed query strategy. Figure \ref{fig:comp_query_stra} provides a visual comparison of the four strategies.

\textbf{Learnable query} is the most commonly used strategy in object detection and segmentation literature~\cite{wang2022omni,zou2023generalized, cheng2021per}. With this strategy, a single set of object queries is trained from scratch, to interact with the image features to encode object location and class information (illustrated in Figure \ref{fig:comp_query_stra} (a)). While the learnable query strategy has been widely used, several works~\cite{zhu2020deformable,zou2023generalized} have shown that its performance on instance/panoptic segmentation and object detection is worse than the conditional query strategy, as discussed in the next section.

\textbf{Conditional query} is proposed in \cite{zhu2020deformable} in order to simulate the proposal generation mechanism present in traditional two-stage object detection frameworks~\cite{ren2015faster}, but adapted for transformer-based detectors. Unlike the learnable query strategy, the conditional query is derived from the segmentation encoder rather than an independent learnable query, as in Figure \ref{fig:comp_query_stra} (b). The segmentation encoder is specifically trained to predict region proposals, from which high-confidence proposals are selected and subsequently input to the segmentation decoder as object queries for final predictions.
The conditional query is more closely aligned with potential objects present in an image, and has consistently demonstrated superior performance for object detection and segmentation tasks~\cite{zhu2020deformable,li2023mask}. However, it is important to note that not all segmentation tasks benefit from this conditional query strategy. For instance, in semantic segmentation, there is often a stuff class that typically corresponds to background regions without defined spatial extent or shape. Conditional queries derived from image features may not effectively capture the characteristics of such background regions, resulting in suboptimal results~\cite{zhang2023simple}. 

\textbf{Separated query} is proposed in OpenSeeD~\cite{zhang2023simple} to handle the limitation of the conditional query strategy on stuff class, in which the query set is divided into learnable and conditional queries, and the training data is also divided into stuff classes and thing classes beforehand.
The learnable and conditional queries are associated with the stuff and thing classes, respectively. After grouping, these two sets of queries are treated independently during the loss computation, and Hungarian matching is applied within each set and there is no interaction. An illustration is given in Figure \ref{fig:comp_query_stra} (c). This separated query strategy effectively resolves the previously encountered issues where the learnable query strategy underperforms in instance segmentation, while the conditional query strategy struggles with stuff class. However, it introduces a few challenges. First, clear separation between stuff and thing classes is required, but, in definition, there is no clear boundary between stuff and thing classes. For example, ``window'' and ``table'' classes are labeled as thing in ADE20K~\cite{zhou2017scene} but as stuff in COCO~\cite{lin2014microsoft}. This will lead to inconsistency and confusion during multiple dataset training.
Second, explicit stuff/things annotations are not always available on public datasets. In fact, expect for small-scale COCO and ADE20K, many datasets lack such annotations \eg large-scale Objects365~\cite{shao2019objects365}, OpenImages~\cite{kuznetsova2020open} etc. Some other datasets, \eg Visual Genome~\cite{krishna2017visual} even can not classify their open-vocabulary objects into stuff/things. Eventually,
separated query can not leverage these datasets for training at all. Third, at inference, especially in open-vocabulary scenarios, inherent errors may arise. In open-vocabulary settings, some testing classes of interest may not be seen during training, making it unclear whether it should be categorized as stuff or thing and which query set to use. For example, consider the case of ``a pile of books on the table''. Even for human, it is still confusing to consider it as stuff or thing.
If one intends to employ OpenSeeD for open-vocabulary segmentation, a preceding binary classification of textual queries into stuff or thing classes would be necessary, either through heuristic assignment or a trained classifier. Nevertheless, both approaches may introduce bias and errors.

\textbf{Mixed query} To address these problems in existing three query strategies, we develop a novel query strategy, mixed query, as shown in Figure \ref{fig:comp_query_stra} (d). Similar to separated query, mixed object query set also consists of two sets of queries, learnable and conditional. However, the mixed query strategy does not impose distinctions based on classes, tasks, or loss computations. The Hungarian matching is carried out between all object queries and all ground truth objects without thing/stuff discrimination. This design enables three advancements. First, mixed query enhances adaptability due to the integration of the dynamic query selection design. This design allows different tasks or examples to dynamically choose the most suitable queries without requiring rigid assignments, thereby avoiding the issues of distinguishing stuff from thing classes of the separated query strategy. For example, some certain classes, such as ``tree'' and ``road'', annotated as stuff in datasets like COCO, are not necessarily background regions and can benefit from the use of conditional queries. Conversely, some other classes, despite falling under thing classes like ``car'', may be closer to be background class for the situations like ``a person sitting inside a car''. In such scenarios, learnable queries are more appropriate to represent these objects. Some typical examples are shown in Figure \ref{fig:stuff_thing_query}. When compared to the separated query strategy, the mixed query exhibits greater adaptability. Second, mixed query eliminates dependence on the stuff/thing classes annotation during training. This flexibility potentially facilitates leveraging more existing datasets that lack such annotations for training purposes. Third, mixed query eliminates query selection errors at inference. Hungarian matching and loss computation are performed on the query embedding outputs of both query sets together. This allows the extraction of the most confident predictions during inference without the necessity to choose a specific query set, thereby averting query set selection errors. Benefiting from these advantages, mixed query strategy enables us to unify the training and inference of various segmentation tasks and datasets within one single model, which will be discussed in the next section.

\subsection{Unified Segmentation Training}

The MQ-Former is trained with the training set $\mathcal{D} = \{(\mathbf{x}_i, \mathbf{y}_i)\}^N_{i=1}$ where $\mathbf{x}_i$ is the image and $\mathbf{y}_i = \{(c_j, \mathbf{b}_j, \mathbf{m}_j)\}^B_{j=1}$ its $B$ annotations. $(c_j, \mathbf{b}_j, \mathbf{m}_j)$ is a triplet depicts a single mask annotation on the image. $c_j$ is the semantic class label of thing or stuff, \eg, ``apple'', ``road'', or a text description, \eg, ``a person wearing a red shirt'', to describe the semantic information characterized with the binary mask region $\mathbf{m}_j$. $\mathbf{b}_j$ is the bounding box (bbox) annotation of this region. 
The model is trained with loss function as follows (for clarity, we omit the weight for each loss term),
\begin{equation}
\begin{split}
    \mathcal{L} = &\sum_{(\mathbf{x}_i, \mathbf{y}_i) \in \mathcal{D}} \sum_{(c_j, \mathbf{b}_j, \mathbf{m}_j) \in \mathbf{y}_i} \mathcal{L}_c (\mathbf{P}^c(\mathbf{x}_i), \mathbf{H}(c_j)) \\
     + & \mathcal{L}_b(\mathbf{P}^b(\mathbf{x}_i),\mathbf{b}_j) + \mathcal{L}_m(\mathbf{P}^m(\mathbf{x}_i),\mathbf{m}_j),
\end{split}
  \label{eq:unified_loss}
\end{equation}
where $\mathcal{L}_c$, $\mathcal{L}_b$, $\mathcal{L}_m$ are the class, bbox and mask loss, respectively. 
They are applied to class, bbox and segment mask embeddings, $\mathbf{P}^c, \mathbf{P}^b, \mathbf{P}^m$, from the decoder outputs and text embedding $\mathbf{H}$, for supervision.
The class loss is the focal loss \cite{lin2017focal} applied on the dot-product between the class embedding and text embedding. The bbox loss is generalized IoU and L1 loss~\cite{rezatofighi2019generalized} between the bounding box embedding and ground truth. The mask loss is calculated with generalized dice loss~\cite{sudre2017generalised} on the mask prediction which is derived from the mask embedding and a pixel encoder. 

Next, we introduce how all segmentation tasks and datasets can be formulated and trained altogether. The integration for instance/semantic/panoptic segmentation is straightforward, because each instance or semantic region has a class label and a mask region annotation. For foreground segmentation, it is regarded as a binary segmentation problem, with the semantic labels of $\{``\text{foreground}",\\``\text{background}"\}$. For referring segmentation, the class label is a textual description. All semantic class labels are in the form of textual description, and will be encoded by the text encoder, as shown in Figure \ref{fig:arch}. So the model is capable of dealing open-vocabulary scenarios and there is no need for sophisticated label space alignment across datasets with different semantic labels.
For all these tasks, the bbox annotation can be derived from the mask annotation. 
Since all datasets are converted to the same annotation format, it is straightforward to merge all datasets together during training.

\subsection{Enhancement with Synthetic Data}

\begin{figure}[t]
\setlength{\abovecaptionskip}{-2.0pt}
\setlength{\tabcolsep}{2pt}
\begin{center}
  \includegraphics[width=1.0\linewidth]{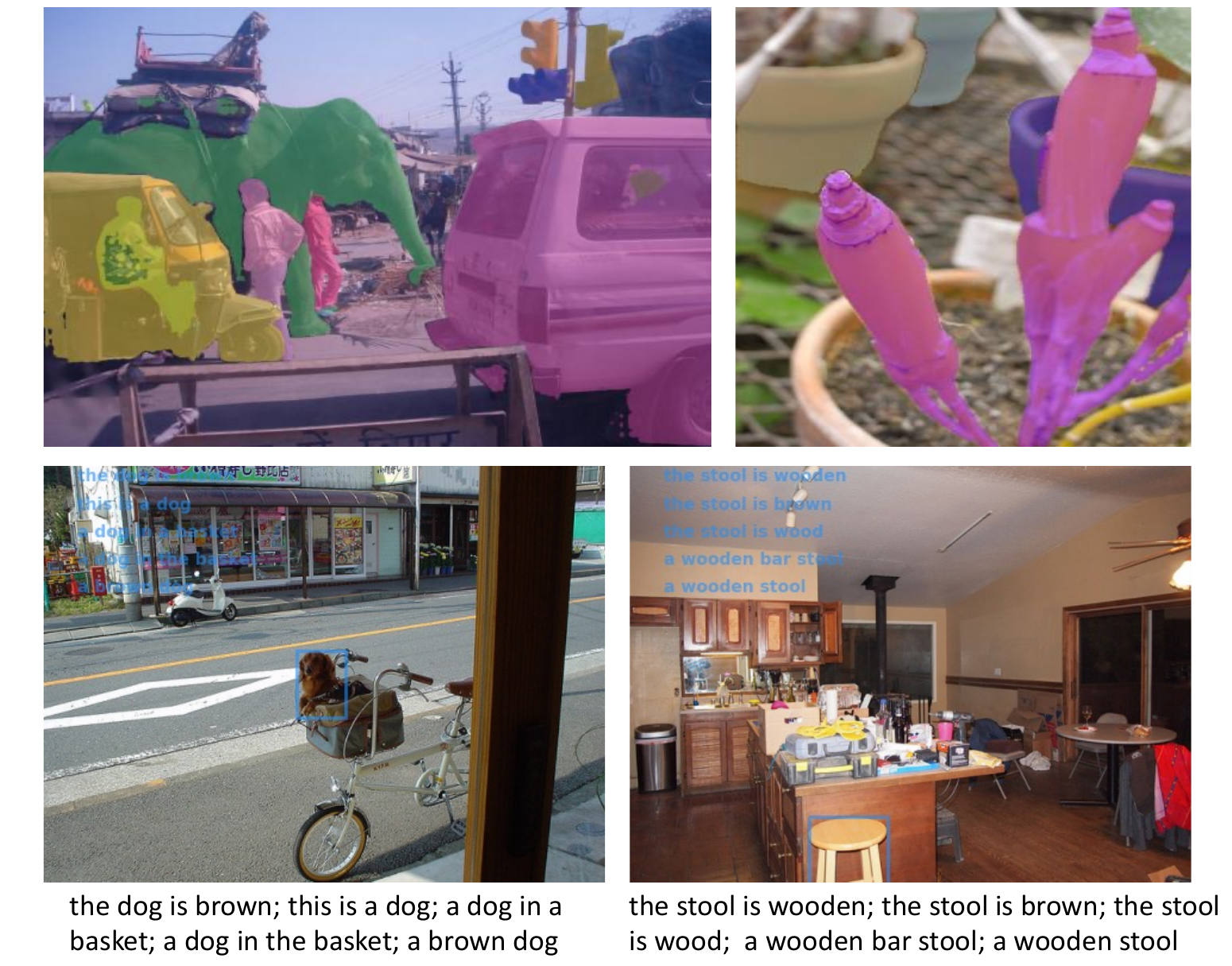}
\end{center}
  \caption{Synthetic data visualization. Upper: synthetic masks by SAM; Lower: synthetic captions by OFA-akin model.}
\label{fig:syn_data}
\end{figure}

Due to the property of multi-task, multi-dataset training, MQ-Former potentially can leverage datasets of different tasks for improvement as many as possible. However, the sizes of well curated segmentation datasets are usually relatively small \footnote{Although SA-1B \cite{kirillov2023segment} is large, it relies on machine predictions and does not have semantic labels.} because pixel-wise mask annotation is expensive, which is not enough to train a strong universal model that can handle all use cases. To circumvent this challenge, we propose to use synthetic data, which has been proven effective to strengthen the model, for instance, in object detection~\cite{cho2023open,gao2022open} image captioning~\cite{davide2023synthcap}. Some recent models can generate high-quality synthetic segmentation masks (e.g SAM~\cite{kirillov2023segment}) and synthetic captions (\eg, OFA \cite{wang2022ofa}, BLIP-2~\cite{li2023blip}), as shown in Figure \ref{fig:syn_data}. In this work, we intend to use those kinds of synthetic data for improving model segmentation performance. 
Since the synthetic data is cheap to generate, it is easy to scale up to a large training set for our MQ-Former training. 

\textbf{Synthetic segmentation mask}:  Instead of generating synthetic segmentation masks directly on unlabeled image, it is a much easier task to segment the mask given an object bounding box because some recent works have shown that they are pretty good at this task~\cite{kirillov2023segment,ke2023segment,zou2023segment}. The size of object detection dataset is usually more than dozen times larger than that of segmentation, \eg, Objects365~\cite{shao2019objects365} of 1.7M images v.s. COCO~\cite{lin2014microsoft} of 120K images. With the generated synthetic masks, we can convert every object detection dataset to a segmentation dataset to have more diverse training data.

\textbf{Synthetic segmentation caption}: The standard segmentation/detection datasets usually lack rich textual descriptions, \eg, 80 fixed category names for COCO. This is a big challenge for our open-vocabulary segmentation model, especially for the task of referring segmentation. The widely used referring segmentation datasets are RefCOCO, RefCOCO+ and RefCOCOg as well as RefClef~\cite{kazemzadeh2014referitgame,yu2016modeling}, whose combination has only about 40K images. The reason to this small dataset size is because annotating a caption description to every individual object segment is expensive.
In order to enrich the semantic information of the training data and improve the generalization ability of the model, we train a OFA-akin~\cite{wang2022ofa} model on the task of object captioning \cite{krishna2017visual}, \ie, generating synthetic caption for each object given the bounding box. With this object captioning model, we generate five synthetic captions with the highest confidences for each object, and use them to expand the training data size. One of the synthetic captions is randomly selected per object at each training iteration.
Our experiments show that jointly training with such synthetic data can largely enrich the semantic information of the training data and improve the generalization ability.

\section{Experiments}

Because MQ-Former is a unified architecture on multiple tasks and datasets, we experiment on a variety of datasets proposed for different tasks: COCO~\cite{lin2014microsoft} and ADE20K~\cite{zhou2017scene} for comprehensive semantic/instance/panoptic annotations; LVIS~\cite{gupta2019lvis} for instance segmentation; RefCOCO, RefCOCO+, RefCOCOg~\cite{kazemzadeh2014referitgame,yu2016modeling} for referring segmentation; HRSOD~\cite{Zeng_2019_ICCV}, DIS~\cite{qin2022}, and other five datasets~\cite{ChengPAMI,mansilla2016oriented,liew2021deep,xie2022pyramid,wang2017learning} for foreground segmentation; Objects365~\cite{shao2019objects365} and Visual Genome~\cite{krishna2017visual} for object detection. In addition, Pascal Context~\cite{mottaghi2014role} and BDD~\cite{yu2018bdd100k} are used for open-set evaluation. 
SeginW benchmark which has 25 datasets is used for open-vocabulary in-the-wild instance segmentation evaluation~\cite{zou2023generalized}. We use mIoU as the evaluation metric for semantic and referring segmentation, Mask AP for instance segmentation, PQ~\cite{kirillov2019panoptic} for panoptic segmentation and MSE for foreground segmentation,
following~\cite{li2023mask,zou2023generalized,zhang2023simple,kim2022revisiting}.
The hyperparameters of the architecture and training follow Mask DINO~\cite{li2023mask}. The pretrained Swin Transformer~\cite{liu2021swin} and CLIP language encoder~\cite{radford2021learning} are adopt as the vision and text encoder, respectively. Swin-Base and CLIP-Base are used as default in most of our experiments due to limited resources, except in Section \ref{sec: sota} where the larger-scale variants are used for the comparison with the state-of-the-art. It is noted that any vision or language backbone encoders can be used by MQ-Former. 
The mixed query set consists of 100 learnable and 300 conditional queries, following some popular settings, e.g. Mask DINO~\cite{li2023mask} (100 learnable queries) and OpenSeeD~\cite{zhang2023simple} (300 queries for things classes). For more details, please refer to the supplementary materials.

\subsection{Comparison among Different Query Strategies}

\begin{table}[t]
\setlength{\abovecaptionskip}{-1.0pt}
\setlength{\tabcolsep}{2pt}
\scriptsize
\begin{center}
    \caption{The performance comparison of different query strategies.}\label{tab:abl_query_strategy}
    \begin{NiceTabular}{l|c|cc|c|c}
\toprule
Query&\multirow{2}{*}{Training data} & \multicolumn{2}{c}{COCO} & ADE&SeginW\\
strategy& &Mask AP &PQ & mIoU &Mask AP\\
\hline
\multirow{2}{*}{learnable}&COCO pano+ADE sem &48.1 &54.3  &50.4 &27.8\\
& COCO pano+ADE sem+VG+refer& 48.6& 54.1 &50.1&32.1\\
\hline
\multirow{2}{*}{conditional}&COCO pano+ADE sem &49.8 &56.5  & 43.2&29.4\\
& COCO pano+ADE sem+VG+refer&49.5&56.2  & 43.9&34.7\\
\hline
separated&COCO pano+ADE sem & 49.6 & 56.3&50.1 &30.0\\
\hline
\multirow{2}{*}{mixed}& COCO pano+ADE sem&49.6 & 56.5 &51.7&30.6\\
& COCO pano+ADE sem+VG+refer& 49.9&  56.8&52.1 &38.4\\
\bottomrule
\end{NiceTabular}
\end{center}
\end{table}

Our initial series of experiments is designed to substantiate the multi-task multi-dataset training efficacy of the proposed mixed query. The basic training set comprises COCO with panoptic annotations (``COCO pano''), providing a common ground for all query strategies detailed in Table \ref{tab:abl_query_strategy}. In addition to COCO, ADE20K with semantic segmentation annotations (``ADE sem''), of different dataset and task, is included in the training set. Visual Genome with instance segmentation (``VG'') and referring segmentation RefCOCO/RefCOCO+/RefCOCOg (``refer'') datasets are also incorporated, with the exception for the separated query strategy as it can not support them. The evaluation uses ADE and COCO for closed-set performance, while SeginW is utilized to assess the open-set generalization capabilities of the models.

The mixed-query is compared to other strategies with the same total query number of 400. The comparison reveals that, in contrast to the learnable query, all other strategies exhibit advancements in instance segmentation tasks, with the mixed query demonstrating the most superior performance. Notably, a considerable degradation is observed in the performance of the conditional query on ADE (mIoU) semantic segmentation, a phenomenon previously discussed and in alignment with observations made in \cite{zou2023generalized,zhang2023simple}. The separated query strategy successfully mitigates issues present in learnable and conditional queries. However, owing to its reliance on stuff/things annotations, its application is limited, hindering its extension to broader training sets and resulting in suboptimal performance on open-set SeginW evaluations. Importantly, the mixed-query not only exhibits comparable performance to the separated query in closed-set evaluations but also overcomes the challenges associated with thing/stuff class differentiation. This innovative aspect enables its training on more diverse datasets, consequently enhancing the model's generalization ability, as evidenced by the significant performance gains on SeginW.

\begin{figure}[t]
\setlength{\abovecaptionskip}{-2.0pt}
\setlength{\tabcolsep}{2pt}
\begin{center}
  \includegraphics[width=1.0\linewidth]{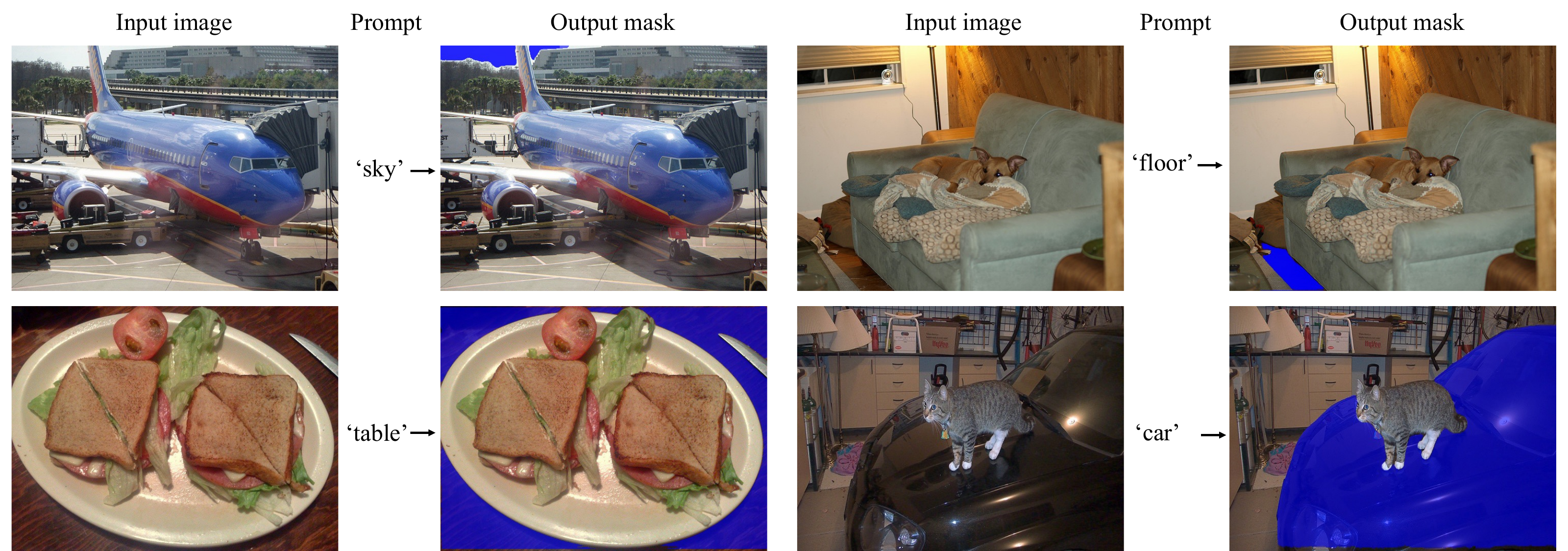}
\end{center}
  \caption{The benefits of dynamic query selection of mixed query strategy. Upper: the stuff objects are predicted with conditional queries instead of learnable queries; Lower: the thing objects are predicted with learnable queries instead of conditional queries.}
\label{fig:stuff_thing_query}
\end{figure}

The superior behavior of mixed query benefits from its dynamic query selection. Unlike separated query that imposes a heuristic association between learnable (conditional) queries and stuff (thing) class, mixed query strategy removes this constraint, evidenced by some supporting experimental results below. To verify this, we check the ratio of thing (stuff) objects that are predicted with the conditional (learnable) queries, which is $99.6\%$ ($53.3\%$) on COCO panoptic segmentation. This means that almost all thing object predictions come from conditional queries, around half of the stuff objects are predicted with the learnable queries (the other half with the conditional ones). 
A similar trend is observed on ADE panoptic segmentation, with ratios of $99.8\%$ and $61.4\%$ respectively.
The results indicate that two types of queries are indeed selected dynamically, supporting the motivation of mixed query strategy. Figure \ref{fig:stuff_thing_query} shows some examples that have ``counter'' assignment under mixed query mechanism. We can see that, sometimes, it is not the optimal solution to strictly associate the thing/stuff objects to conditional/learnable queries. For example, the object of table (bottom-left of Figure \ref{fig:stuff_thing_query}) is closer to stuff than thing, and it should not have clear discrimination between thing and stuff.

\subsection{Enhancement by Synthetic Data}

\begin{table}
\setlength{\abovecaptionskip}{-1.0pt}
\setlength{\tabcolsep}{3pt}
\small
\begin{center}
\caption{The impact of synthetic masks}\label{tab:abl_syn_mask}
\begin{NiceTabular}{l|c|c}
\toprule
Training data & Mask AP & Box AP \\
\hline
COCO ins & 49.7 & 55.3 \\
COCO ins + syn-Objects365 & 50.5 & 56.8 \\
\bottomrule
\end{NiceTabular}
\end{center}
\end{table}

\begin{table}
\setlength{\abovecaptionskip}{-1.0pt}
\setlength{\tabcolsep}{3pt}
\small
\begin{center}
\caption{The impact of synthetic captions}\label{tab:abl_syn_cap}
\begin{NiceTabular}{l|c}
\toprule
Training data & mIoU \\
\hline
RefCOCOg & 57.8 \\
syn-COCO & 58.8 \\
RefCOCOg + syn-COCO & 62.6 \\
\bottomrule
\end{NiceTabular}
\end{center}
\end{table}

In order to validate the benefits of synthetic masks, $30\%$ images are sampled from Objects365 \cite{shao2019objects365} training set and synthetic mask is generated for each object with \cite{ke2023segment}. This subset is denoted as syn-Objects365. We jointly train a model on COCO with instance annotation (``COCO ins'') and syn-Objects365 and compare with the baseline trained on ``COCO ins'' only. As shown in Table \ref{tab:abl_syn_mask}, the improvement is clear, suggesting the benefit of using synthetic masks.

Similarly, synthetic object captions are generated for all COCO instances, denoted as syn-COCO. We trained a model jointly on it with RefCOCOg. The comparison in Table \ref{tab:abl_syn_cap} with the baseline shows that the improvement is significant (more than 4 points), indicating the benefits of synthetic captions.


\subsection{Enhancement by Scaling up Datasets and Tasks}

\begin{table*}[t]
\setlength{\abovecaptionskip}{-1.0pt}
\setlength{\belowcaptionskip}{-1.0pt}
\scriptsize
\begin{center}
\caption{The performance of MQ-Former trained on more tasks and datasets.}\label{tab:scale_up}
\begin{NiceTabular}{l|cccc|c|c|c}
\toprule
\multirow{2}{*}{Training data} & \multicolumn{4}{c|}{COCO} & UHRSD &RefCOCOg&SeginW\\
& PQ & Mask AP & Box AP & mIoU & MSE & mIoU&Mask AP\\
\hline
COCO pano+ADE pano & 56.5 & 48.5 &54.2 & 65.3 & 0.3& 31.7& 30.0\\
COCO pano+ADE pano+fore & 56.3 & 48.3 &54.3 & 65.1 & 0.03& 32.1 &30.4\\
COCO pano+ADE pano+fore+refer & 56.0 & 48.5 &54.5 & 65.0 & 0.03&63.4&31.2\\
COCO pano+ADE pano+fore+refer+ins$^+$& 56.8 & 49.1 & 55.1& 65.4 & 0.03& 64.3& 38.6\\
\bottomrule
\end{NiceTabular}
\end{center}
\end{table*}

MQ-Former can be jointly trained beyond the generic instance/semantic/panoptic segmentation tasks and datasets to further improve the model generalization ability. In this section, we incorporate more datasets and tasks for joint training. Starting from the combination of COCO and ADE panoptic annotation (``COCO pano + ADE pano'') setting, we train the model with more and more tasks and datasets, and report the performance on each benchmark in Table \ref{tab:scale_up}, where ``+fore'' means adding additional seven foreground segmentation datasets, ``+refer'' adding additional referring segmentation of RefCOCO, RefCOCO+, RefCOCOg and syn-COCO, and ``+ins$^+$'' adding additional instance segmentation of LVIS, Visual Genome and syn-Objects365. COCO, UHRSD, RefCOCOg and SeginW are used as for panoptic/instance/semantic, foreground, referring, and open-vocabulary segmentation evaluation, respectively.
It can be seen that adding new datasets or tasks does not hurt the performance on the original tasks and datasets, but the model can handle new tasks or data domains, \eg, the huge improvement on UHRSD when adding foreground segmentation and on RefCOCOg when adding referring segmentation for training.
If the new task is similar to the early ones, it could have some improvement on the early tasks. For example, when jointly training with larger and more diverse instance segmentation datasets (``+ins$^+$''), the instance segmentation performance on COCO is improved by 0.6 points. 
The most benefit of jointly training is that when more diverse data is used, the generalization ability of the model will be improved. This is evidenced by the significant performance improvement on SeginW from 30.0 to 38.6.

\begin{table*}[h]
  \centering
  \setlength{\tabcolsep}{2pt}
  \scriptsize
    \caption{The comparison to state of the arts on open-set benchmarks (left) and closed-set benchmarks (right). `$\times$' represents that the method is not capable of handling the task; `$-$' represents no results reported in the original paper; `$\star$' represents our implementation based on the published codes. We bold the best entry and underline the runner-up in each column. (T)/(L) means tiny/large model scale.}
  \label{tab:sota}
  \begin{tabular}{ccccc}
    \begin{NiceTabular}{l|cccc|cc|c|c}
\toprule
\multicolumn{9}{c}{Open-set comparison}\\
\hline
 \multirow{2}{*}{} & \multicolumn{4}{c|}{ADE}&PC-59&PC-459&BDD&SeginW\\
Method & PQ & Mask AP& Box AP & mIoU &mIoU & mIoU & PQ&Mask AP\\
\hline
LSeg+~\cite{ghiasi2022scaling} & -& -& -& 18.0 &46.5&7.8&-&-\\
MSeg~\cite{lambert2020mseg} & - & - & - & 19.1 & - & - & - & -\\
ZegFormer~\cite{ding2022decoupling} & - & - & - & 16.4 & - & - & - & -\\
SPNet~\cite{xian2019semantic} & - & - & - & - & 24.3 & - & - & -\\
DaTaSeg~\cite{gu2024dataseg}& -& -& -& - &51.4&11.1&-&-\\
MaskCLIP~\cite{ding2022open} & 15.1& 6.0& -& 23.7 &45.9&10.0&-&-\\
GroupViT~\cite{xu2022groupvit}& -& -& -& 10.6 &25.9&4.9&-&-\\
OpenSeg~\cite{ghiasi2022scaling} & -& -& -& 21.1 &42.1&9.0&-&-\\
ODISE~\cite{xu2023open} & {\bf 22.6}& 14.4& 15.8& {\bf29.9} &57.3&14.5&-&-\\
X-Decoder~\cite{zou2023generalized} & 21.8 & 13.1&- & 29.6 & \underline{64.0} & \underline{16.1} & 17.8 & 32.3 \\
OpenSeeD~\cite{zhang2023simple} & 19.7& \underline{15.0}& \underline{17.7}& 23.4 &-&-&\underline{19.4}&\underline{36.1}\\
\hline
MQ-Former (L) & \underline{22.1}& {\bf17.3}& {\bf19.2}& \underline{25.0} &{\bf65.0}&{\bf18.1}&{\bf29.3}&{\bf 43.4}\\
\bottomrule
\end{NiceTabular}&
    \begin{NiceTabular}{l|cccc|c|c|c|c}
\toprule
\multicolumn{9}{c}{Closed-set comparison}\\
\hline
 \multirow{2}{*}{} & \multicolumn{4}{c|}{COCO} & \multicolumn{2}{c|}{ADE}&RefCOCOg&UHRSD\\
Method & PQ & Mask AP& Box AP & mIoU & PQ & mIoU &mIoU & MSE\\
\hline
LAVT~\cite{yang2022lavt} & $\times$& $\times$& $\times$&$\times$ &$\times$ &$\times$&63.3 &$\times$\\
PolyFormer~\cite{liu2023polyformer} & $\times$& $\times$&$\times$ &$\times$ &$\times$ &$\times$&{\bf71.2} &$\times$\\
PGNet~\cite{wang2021pgnet} & $\times$& $\times$& $\times$&$\times$ &$\times$ &$\times$&$\times$&0.04\\
InSPyReNet~\cite{kim2022revisiting} & $\times$&$\times$ &$\times$&$\times$ &$\times$ &$\times$&$\times$&{\bf0.02}\\
Mask2Former~\cite{cheng2022masked} & 57.8 &48.6 &52.1 &67.4&48.1&56.1&$\times$&$\times$\\
Mask DINO~\cite{li2023mask} &58.3 & 50.6 &56.2 &67.5&-& - &$\times$&$\times$\\
OneFormer~\cite{jain2023oneformer} & 57.9 &49.0 &- &67.4&51.4&  57.0 &$\times$&$\times$\\
X-Decoder~\cite{zou2023generalized} & 56.9 &46.7 &- &67.5&49.6& 58.1 &64.6&$\times$\\
OpenSeeD (T)$^\star$~\cite{zhang2023simple} & 53.4 & 47.1& 51.2& 63.8&-&  - &$\times$&$\times$\\
OpenSeeD (L)~\cite{zhang2023simple} & {\bf 59.5} &{\bf53.2 }&\underline{58.2} &{\bf68.6}&{\bf53.1}&  {\bf58.6} &$\times$&$\times$\\
\hline
MQ-Former (T) & 54.6 & 47.6& 51.6& 63.9&-&  - &$\times$&$\times$\\
MQ-Former (L)& \underline{58.8} &\underline{52.3} &{\bf58.4} &\underline{68.4}&\underline{52.6}& \underline{58.1} &\underline{68.2}&\underline{0.03}\\
\bottomrule
\end{NiceTabular}
  \end{tabular}
\end{table*}

\subsection{Comparison with the state-of-the-art}
\label{sec: sota}

Finally, we scale up the visual and text encoders to Swin-Large and CLIP-Large for state-of-the-art (SOTA) comparison. Based on the training set combination of ``COCO pano+ADE pano+fore+refer+ins$^+$'', in order to leverage more training data, we further generate synthetic masks and object captions for full Objects365. The comparison is conducted on various benchmarks and scenarios in Table \ref{tab:sota}.

\noindent\textbf{Open-set segmentation} Open-set evaluation stands as a critical metric for assessing the generalization ability of a model, providing insights into its adaptability and performance in real-world applications. Excluding ADE in training, we evaluate the zero-shot performance on ADE20K for panoptic/semantic/instance segmentation, Pascal Context 59 (PC-59) with 59 common classes and PC-459 with full 459 classes~\cite{mottaghi2014role} for semantic segmentation, and BDD~\cite{yu2018bdd100k} for panoptic segmentation. The results are presented in Table \ref{tab:sota} (left). MQ-Former improves the state-of-the-art open-set instance segmentation on ADE by 2.3, semantic segmentation on PC-59 (-459) by 1.0 (2.0) and panoptic segmentation on BDD by 9.9 percentage. ODISE has higher PQ and mIoU on ADE probably because it uses huge-scale backbones.

\noindent\textbf{Segmentation in the wild} In order to further evaluate the generalization ability of MQ-Former, we evaluate it on the in-the-wild benchmark SeginW \cite{zou2023generalized}. This benchmark contains 25 domain-specific datasets for instance segmentation tasks. The evaluation is conducted under the zero-shot setting. The comparison results are given in Table \ref{tab:sota} (left, and the last column), where our model has a significant improvement (7.3 points) over the prior art. This benefits from the joint datasets training, which includes more training data and classes, enabling a model of stronger generalization.

\noindent\textbf{Closed-set segmentation} Finally, we evaluate on closed-set segmentation through a comparison across five tasks, \ie, instance, semantic, panoptic, referring and foreground segmentation  
as listed in Table \ref{tab:sota} (right). Two types of SOTA are compared, task or dataset specific models, like PolyFormer~\cite{liu2023polyformer}, InSPyReNet~\cite{kim2022revisiting}, and unified models, like Mask DINO~\cite{li2023mask}, OneFormer~\cite{jain2023oneformer}. 
First, all other models are unable to handle one or a few listed tasks but MQ-Former can cover all. Second, our model can achieve comparable performances to SOTA, using one {\it single} model. Third, on referring segmentation, our model outperforms X-Decoder that has more complex cross-modality fusion decoder architecture. Finally, our model is slightly worse than OpenSeeD in the large-scale model setting. However, OpenSeeD only released its tiny-scale model trained on COCO panoptic annotation, which prevents us from finding where is the gap in the large-scale unified experimental setting. For fair comparison, we train MQ-Former with the same tiny model size and COCO dataset as OpenSeeD, but MQ-Former achieves better results on all COCO evaluation metrics than OpenSeeD. This suggests that the inferiority is not caused by architecture design. More importantly, MQ-Former avoids the heuristic stuff/thing discrimination in OpenSeeD and has the potential of leveraging more diverse data for training. This leads to stronger open-set generalization ability reflected in open-set segmentation.

\section{Conclusion}
\label{sec:conclusion}

In this paper, we propose a unified architecture for various image segmentation tasks. The architecture can be jointly trained across datasets and tasks and applicable to open-vocabulary settings. This benefits from a novel mixed query strategy which dynamically associates the mixed queries to the thing/stuff objects. This design simplifies the query matching mechanism for unified models and enables the model to be trained on different tasks and datasets. In order to improve the generalization ability, the synthetic data has been leveraged. As a result, our model, MQ-Former, has achieved impressive results on a multitude of segmentation tasks with a single unified model.

{
    \small
    \bibliographystyle{ieeenat_fullname}
    \bibliography{main}
}


\clearpage

\appendix
In this supplement, we show some other additional experimental results and details that are not present in the main paper due to the page limitation.

\begin{table*}
\setlength{\abovecaptionskip}{-1.0pt}
\setlength{\tabcolsep}{1.0pt}
\caption{Open-set segmentation comparison on the SeginW benchmark. We bold the best entry in each column.}
\scriptsize
\begin{center}
\begin{NiceTabular}{l|cc|ccccccccccccccccccccccccc}
\toprule
\multirow{2}{*}{Model} & \multirow{2}{*}{Med.} & \multirow{2}{*}{Avg.} & Air- & \multirow{2}{*}{Bottles} & Br. & \multirow{2}{*}{Chicken} & \multirow{2}{*}{Cows} & Ele.-& \multirow{2}{*}{Eleph.} & \multirow{2}{*}{Fruits} & \multirow{2}{*}{Gar.} & Gin.-&\multirow{2}{*}{Hand} & Hand-&House-&HH.-&Nut.-&\multirow{2}{*}{Phones}&\multirow{2}{*}{Poles}&\multirow{2}{*}{Puppies}&\multirow{2}{*}{Rail}&Sal.-&\multirow{2}{*}{Stra.}&\multirow{2}{*}{Tablets}&\multirow{2}{*}{Toolkits}&\multirow{2}{*}{Trash}&\multirow{2}{*}{W.M}\\
&  &  &Par. & & Tum. & & &Sha. & & & & Gar. & & Metal & Parts & Items & Squi. & & & & &Fil. & & & & &\\
\hline
X-Decoder~\cite{zou2023generalized}&22.3&32.3&13.1&42.1&2.2&8.6&44.9&7.5&66.0&79.2&33.0&11.6&75.9&42.1&7.0&53.0&68.4&15.6&20.1&59.0&2.3&19.0&67.1&22.5&9.9&22.3&13.8\\
OpenSeeD~\cite{zhang2023simple} &38.7 & 36.1 & 13.1 & 39.7 & 2.1 & 82.9 & 40.9 & 4.7 & 72.9 & 76.4 & 16.9 & 13.6& 92.7 & 38.7 & 1.8& 50.0& 40.0& 7.6& 4.6& 74.6& 1.8& 15.0&82.8& 47.4&15.4&15.3&52.3\\
MQ-Former &{\bf 43.0} & {\bf43.4} & {\bf14.4} & {\bf44.4}& {\bf3.3}&{\bf 85.2}&{\bf 45.0}&{\bf15.0} &{\bf75.2 }&{\bf80.4} &{\bf33.1 }& {\bf20.9}& {\bf94.4}&{\bf44.6 }&{\bf 7.8}&{\bf54.2} &{\bf69.5} &{\bf16.0 }&{\bf24.2} &{\bf 78.0}&{\bf4.4} &{\bf27.8} &{\bf84.5 }&{\bf49.3 }&{\bf23.2} & {\bf35.5}&{\bf59.4}\\
\bottomrule
\end{NiceTabular}
\end{center}
\label{tab:seginw}
\end{table*}

\section{Full Results of SeginW}

In Section 4.4 of the main paper, in order to investigate the generalization ability of MQ-Former for segmentation, we conduct a zero-shot evaluation of our model on the Segmentation in the Wild (SeginW) benchmark~\cite{zou2023generalized}, which comprises 25 datasets, and report the average mAP of all the datasets. In this supplementary material, we report other additional results including median mAP and individual mAP on each dataset. The results detailed in Table \ref{tab:seginw} show the superiority of MQ-Former over X-Decoder~\cite{zou2023generalized} and OpenSeeD~\cite{zhang2023simple} across all datasets. This indicates that the importance of joint training on more diverse datasets and tasks in enhancing the generalization ability of models, a capability unique to MQ-Former.

\begin{table*}
  \centering
  \setlength{\tabcolsep}{2pt}
    \caption{The improvement of multi-dataset multi-task training over dataset-specific training baselines. The number in the bracket is the gain.}
\footnotesize
  \begin{tabular}{ccccc}
    \begin{NiceTabular}{l|c|cccc|cc|c|c|c}
\toprule
 & \multirow{2}{*}{Training data} & \multicolumn{4}{c|}{COCO} & \multicolumn{2}{c|}{ADE}&RefCOCOg&UHRSD&SeginW\\
 & &PQ & Mask AP& Box AP & mIoU & PQ & mIoU &mIoU & MSE&Mask AP\\
\hline
\multirow{4}{*}{baseline} & COCO & 57.1&50.8 &57.0&66.7&-  &-&-&-&30.2\\
& ADE&-&-&- &-&48.2&54.1  &-&-&-\\
& RefCOCO,RefCOCO+,RefCOCOg&-&-&- &-&-& - &62.2&-&-\\
& foreground datasets&-&-&- &-&-&-  &-&0.02&-\\
\hline
MQ-Former& multiple datasets&58.8(+1.7) &52.3(+1.5) &58.4(+1.4) &68.4(+1.7)&52.6(+4.4)& 58.1(+4.0) &68.2(+6.0)&0.03(-0.01)&43.4(+13.2)\\
\bottomrule
\end{NiceTabular}
  \end{tabular}
  \label{tab:baseline}
\end{table*}

\begin{table}
\setlength{\abovecaptionskip}{-1.0pt}
\setlength{\tabcolsep}{3pt}
\caption{The impact of query numbers.}
\small
\begin{center}
\begin{NiceTabular}{l|c|cc}
\toprule
\multirow{2}{*}{$\#$learnable+$\#$conditional} & ADE & \multicolumn{2}{c}{COCO}\\
& mIoU & Mask AP & Box AP \\
\hline
100+300 & 51.7 & 49.6 &54.9\\
300+900 & 52.0 & 50.7 & 57.4\\
\bottomrule
\end{NiceTabular}
\end{center}
\label{tab:abl_query_num}
\end{table}

\begin{table}
\setlength{\abovecaptionskip}{-1.0pt}
\setlength{\tabcolsep}{3pt}
\caption{The model size and speed comparison.}
\small
\begin{center}
\begin{NiceTabular}{l|cc}
\toprule
Method & Params & FPS \\
\hline
OneFormer~\cite{jain2023oneformer} & 219M & 5.6 \\
X-Decoder~\cite{zou2023generalized} & 280M & 6.1 \\
OpenSeeD~\cite{zhang2023simple} & - & - \\
MQ-Former & 286M & 5.1\\
\bottomrule
\end{NiceTabular}
\end{center}
\label{tab:speed}
\end{table}

\section{Qualitative Results}

\begin{figure*}[t]
\setlength{\abovecaptionskip}{-2.0pt}
\setlength{\tabcolsep}{2pt}
\begin{center}
  \includegraphics[width=1.0\linewidth]{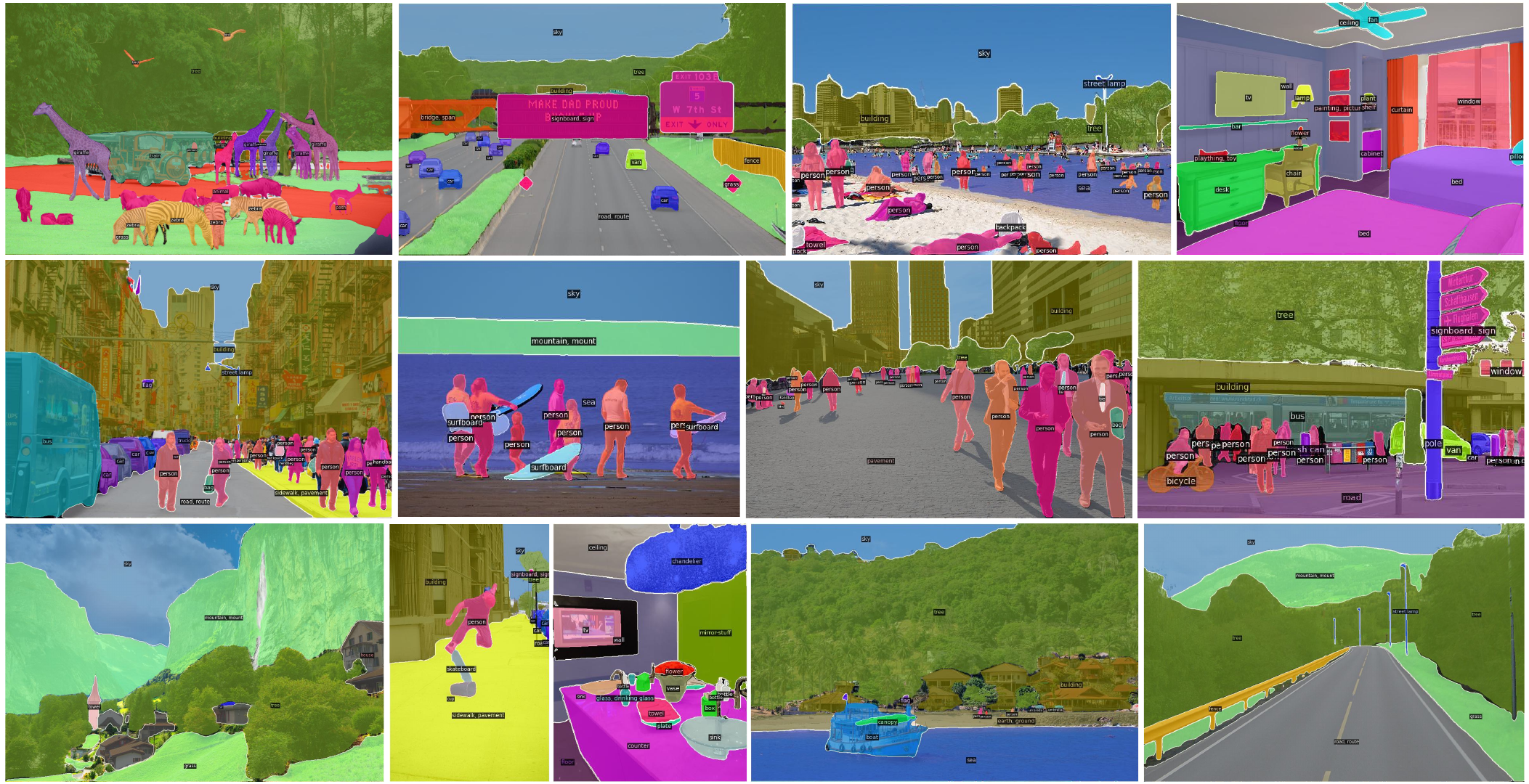}
\end{center}
  \caption{Qualitative visualization on open-set panoptic segmentation.}
\label{fig:vis_pano}
\end{figure*}

\begin{figure*}[t]
\begin{center}
  \includegraphics[width=1.0\linewidth]{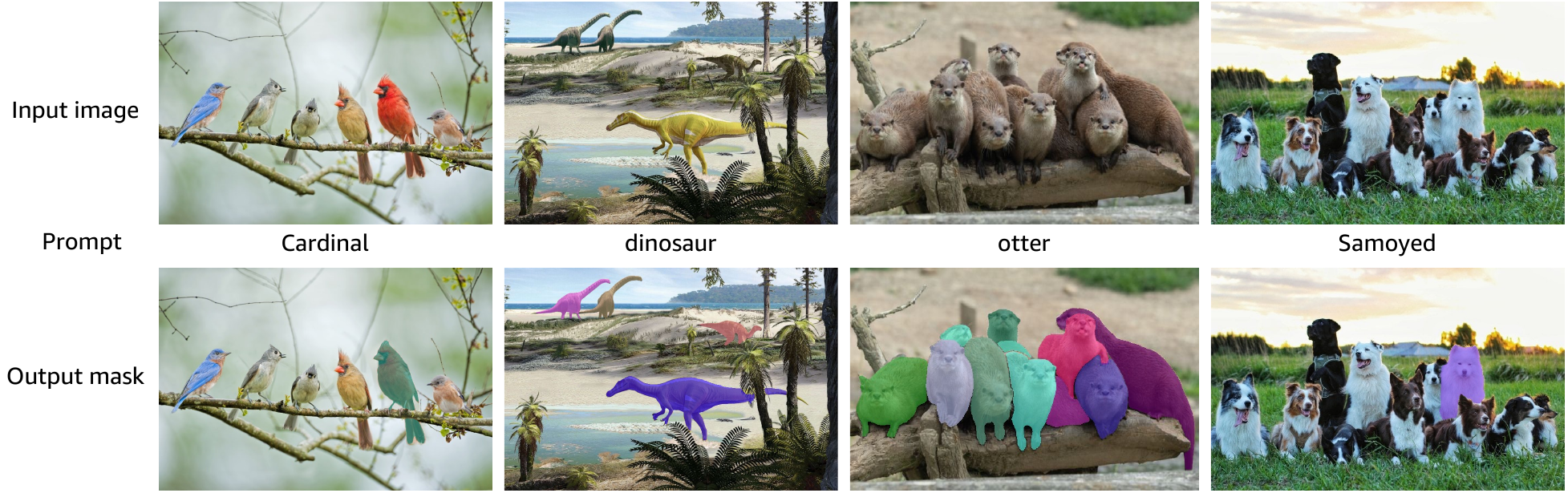}
\end{center}
  \caption{Qualitative visualization on open-set instance segmentation.}
\label{fig:vis_ins}
\end{figure*}

\begin{figure*}[t]
\setlength{\abovecaptionskip}{-2.0pt}
\setlength{\tabcolsep}{2pt}
\begin{center}
  \includegraphics[width=1.0\linewidth]{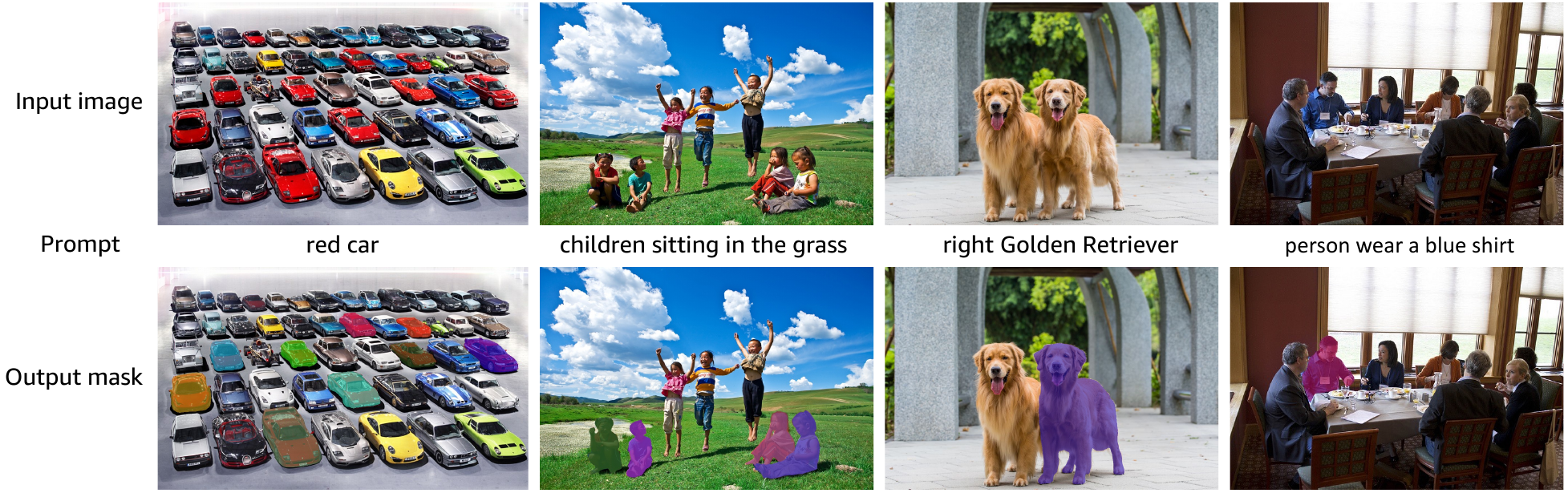}
\end{center}
  \caption{Qualitative visualization on open-set referring segmentation.}
\label{fig:vis_ref}
\end{figure*}

\begin{figure*}[t]
\setlength{\abovecaptionskip}{-2.0pt}
\setlength{\tabcolsep}{2pt}
\begin{center}
  \includegraphics[width=1.0\linewidth]{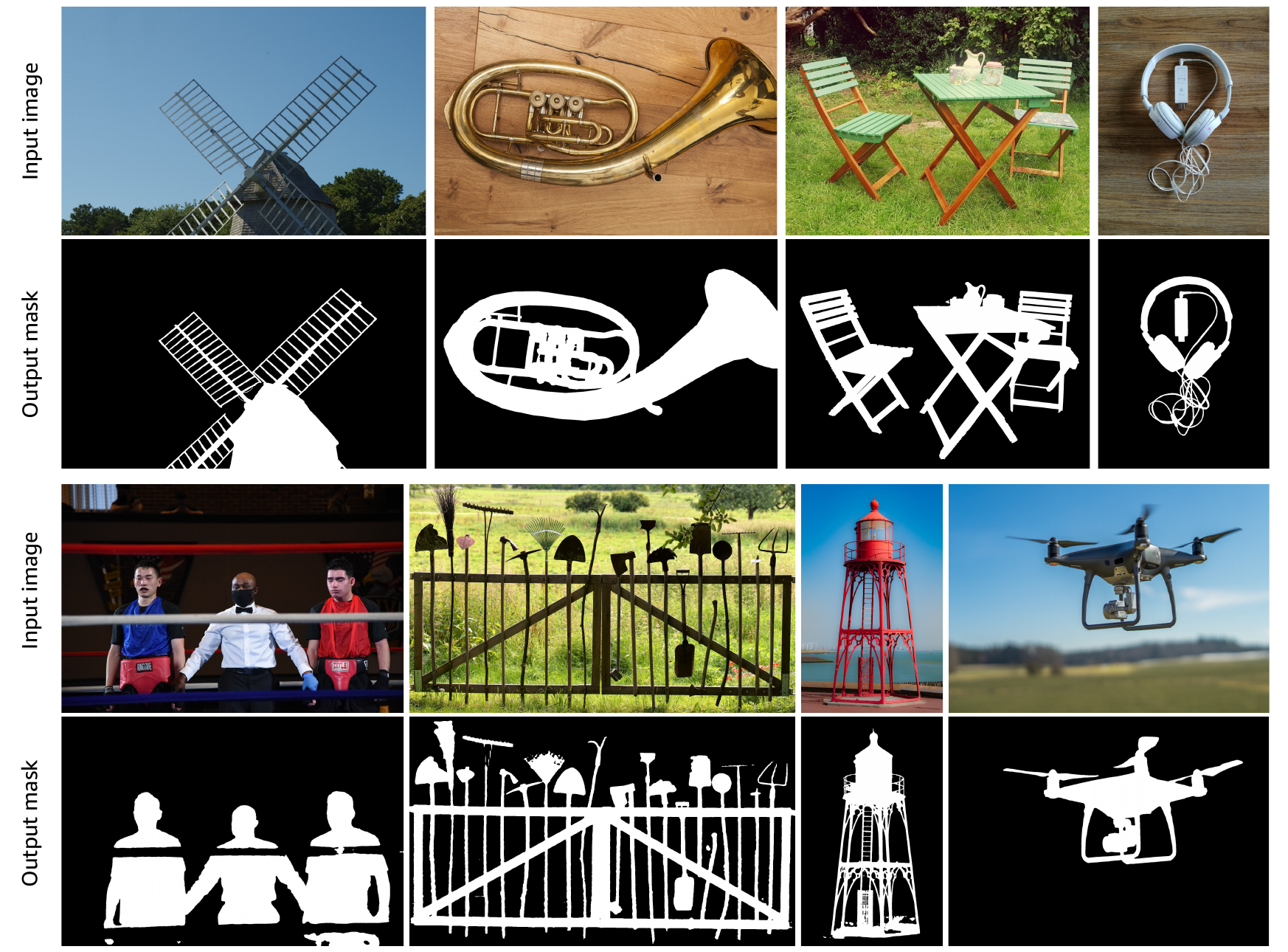}
\end{center}
  \caption{Qualitative visualization on foreground segmentation.}
\label{fig:vis_fore}
\end{figure*}

We present qualitative visualizations for open-set panoptic, instance and referring segmentation in Figures \ref{fig:vis_pano}, \ref{fig:vis_ins}, \ref{fig:vis_ref} and foreground segmentation in Figure \ref{fig:vis_fore}, respectively. The images are randomly selected from the web to provide a diverse and representative set for evaluation.

\section{Baseline for Large-scale Backbones}

\begin{table}
\setlength{\abovecaptionskip}{-1.0pt}
\setlength{\tabcolsep}{3pt}
\caption{Upsampling ratio of joint training. ``referring'' refers to the combination of RefCOCO, RefCOCO+, RefCOCOg~\cite{kazemzadeh2014referitgame,yu2016modeling}. ``foreground'' refers to the combination of seven foreground datasets, HRSOD~\cite{Zeng_2019_ICCV}, DIS~\cite{qin2022}, THUS~\cite{ChengPAMI}, COIFT~\cite{mansilla2016oriented}, ThinObjects5K~\cite{liew2021deep}, UHRSD~\cite{xie2022pyramid}, DUTS~\cite{wang2017learning}.}
\footnotesize
\begin{center}
\begin{NiceTabular}{c|c|c|c}
\toprule
Dataset & Ratio & $\#$Images & $\#$Annotations\\
\hline
COCO & 3 &100K & 1.3M\\
ADE20K & 30 & 20K &271K\\
LVIS & 3 & 100K &1.3M\\
Visual Genome & 9 &100K &2.3M\\
Objects365 & 1 & 1.7M & 25M\\
referring & 6 & 54K &124K\\
syn-COCO & 3 &100K & 1.3M\\
syn-Objects365& 1& 1.7M & 25M\\
foreground & 9 & 100K&100K\\
\bottomrule
\end{NiceTabular}
\end{center}
\label{tab:sampling_ratio}
\end{table}

In Section 4.4 of the main paper, we report the multi-task multi-dataset training results of MQ-Former on each benchmark. In this section, we supply the results of baseline models that are trained on each dataset for reference. In Table \ref{tab:baseline}, we can observe that joint training on multiple datasets and multiple tasks consistently improves the performance on all individual tasks and datasets. These findings underscore the significance of MQ-Former.

\section{The Impact of Query Numbers}

In this section, we ablate the impact of the number of queries. By default, we use mixture of 100 learnable and 300 conditional queries. This setting is derived from MaskDINO, ADE semantic setting of 100 learnable queries and COCO instance setting of 300 conditional queries. It is also the same as OpenSeeD using 100 learnable queries for stuff classes and 300 conditional queries for thing classes. Based on the Base-scale image and text encoder backbones, we train models with the same cross-task cross-dataset configuration, the combination of COCO with instance segmentation and ADE with semantic segmentation, of different queries. In table \ref{tab:abl_query_num}, we observe that increasing the query number can improve the performance.
However, the memory cost also increases considerably. Because such GPU memory cost is not affordable for our team when scaling up to large-scale backbones, in other experiments across the paper, we keep the `100+300' setting consistently. 

\section{Model Size and Speed Comparison}

We evaluate the model size in terms of the numbers of parameters (Params) and conduct a speed comparison by reporting frames-per-second (FPS). The speed tests are performed on A100 NVIDIA GPU with 40GB memory by taking the average computing time with batch size 1 on the entire validation set, using Detectron2~\cite{wu2019detectron2}. All models listed in Table \ref{tab:speed} are characterized by large-scale backbone models. The absence of OpenSeeD~\cite{zhang2023simple} results in the table is because it did not release the checkpoint and full codes. In general, there is no substantial difference in the forward speed across three models. The increase in parameters for both X-Decoder and our MQ-Former over OneFormer is primarily attributed to the introduction of a language encoder, given that they are open-vocabulary models.

\section{Experimental Implementation Details}
\label{sec:details}

\noindent\textbf{Training settings} For the experiments of Sections 4.1 and 4.2, we train our model with a batch size of 32. AdamW is used as the optimizer with a base learning rate of 2e-4 for the segmentation encoder and decoder, and 2e-5 , 10 warmup iterations, and a weight
decay of 0.05. We decay the learning rate at 0.9 and 0.95 fractions of the total number of training steps by a factor of 10. We train for a total of 50 epochs. On the experiments of Section 4.3 and 4.4, the batch size is scaled up to 128.

\noindent\textbf{Datasets} In order to mitigate the data leakage issue, we implement exclusion in our training data. Specifically, for the COCO 2017 training set, examples belonging to RefCOCO, RefCOCO+, RefCOCOg validation sets are excluded. Conversely, training examples from RefCOCO, RefCOCO+, RefCOCOg that overlap with COCO 2017 validation set are also excluded. Similar exclusion procedures are applied to LVIS training set, removing examples associated with the RefCOCO, RefCOCO+, RefCOCOg validation sets. Distinct data augmentation strategies are applied based on the type of training data. For instance, semantic and panoptic data, we follow the augmentation strategy of Mask DINO~\cite{li2023mask}. For referring segmentation data, the augmentation data is the same as instance segmentation but random clip is excluded. For foreground segmentation training data, we follow the data augmentation of InSPyReNet~\cite{kim2022revisiting}. Different upsampling ratios for each dataset are applied during joint training, which are maintained as specified in Table \ref{tab:sampling_ratio}. In total, the MQ-Former is trained on around 2M distinct images examples and 57M mask annotations. It is noted that the comparison in Table 5 is a system-level comparison. The training data varies across each method. For instance, X-decoder~\cite{zou2023generalized} additionally incorporates image-text corpora in its training process.

\section{Ethical Considerations}

We discuss the ethical considerations from three aspects: \textbf{Environmental Impact:} Training MQ-Former requires significant computational resources. The environmental impact of such resource-intensive processes should be taken into account, and efforts should be made to develop more energy-efficient algorithms. \textbf{Transparency and Explainability:} Like other deep learning models, MQ-Former is also considered ``black boxes'' because it is challenging to understand how they reach specific decisions. Ensuring transparency and explainability is essential to build trust and accountability, especially in applications with significant consequences. \textbf{Bias and Fairness:} Like other machine learning models, image segmentation models can be biased based on the data they are trained on. If the training data is not diverse and representative, the model may perform poorly on certain demographics or groups, perpetuating existing biases. However, this problem can be resolved to a certain extent by MQ-Former thanks to its versatility of joint training on multiple diverse datasets and tasks.

\section{Limitations}

Recently, a newly emerging reasoning segmentation task has been introduced~\cite{lai2023lisa}. The task is designed to output a segmentation mask given a complex and implicit query text. For example, given an image with various fruits, the query is ``what is the fruit with the most Vitamin C in this image''. This task demands a level of reasoning typically handled by multi-modal Large Language Models. Currently, MQ-Former does not explicitly support this task. However, addressing this limitation is part of our agenda for future research.

Additionally, in this version of MQ-Former, we have not incorporated cross-modality feature fusion. While Grounding DINO~\cite{liu2023grounding} has demonstrated the effectiveness of this fusion in object detection and it has been successfully applied to X-Decoder~\cite{zou2023generalized} for image segmentation, it's important to mention that introducing cross-modality feature fusion would come at the cost of increased training resources and inference time. Remarkably, even without cross-modality feature fusion, MQ-Former has exhibited superior performance compared to X-Decoder, as evident in Table 6 of the main paper. This indicates the importance of our approach, emphasizing the benefit of training on multiple datasets and tasks.

\end{document}